# Rule of Three for Superresolution of Still Images with Applications to Compression and Denoising

**Mario Mastriani**

DLQS LLC, 4431 NW 63RD Drive, Coconut Creek, FL 33073, USA.
mmastri@gmail.com

*Abstract*—We describe a new method for superresolution of still images (in the wavelet domain) based on the reconstruction of missing details subbands pixels at a given *ith* level via Rule of Three (Ro3) between pixels of approximation subband of such level, and pixels of approximation and detail subbands of (*i+1*)*th* level. The histogramic profiles demonstrate that Ro3 is the appropriate mechanism to recover missing detail subband pixels in these cases. Besides, with the elimination of the details subbands pixels (in an eventual compression scheme), we obtain a bigger compression rate. Experimental results demonstrate that our approach compares favorably to more typical methods of denoising and compression in wavelet domain. Our method does not compress, but facilitates the action of the real compressor, in our case, Joint Photographic Experts Group (JPEG) and JPEg2000, that is, Ro3 acts as a catalyst compression.

## I. INTRODUCTION

The central aim of Super-Resolution (SR) is to generate a higher resolution image from lower resolution images. High resolution image offers a high pixel density and thereby more details about the original scene. The need for high resolution is common in computer vision applications for better performance in pattern recognition and analysis of images. High resolution is of importance in medical imaging for diagnosis. Many applications require zooming of a specific area of interest in the image wherein high resolution becomes essential, e.g. surveillance, forensic and satellite imaging applications [1]. However, high resolution images are not always available. This is since the setup for high resolution imaging proves expensive and also it may not always be feasible due to the inherent limitations of the sensor, optics manufacturing technology. These problems can be overcome through the use of image processing algorithms, which are relatively inexpensive, giving rise to concept of super-resolution. It provides an advantage as it may cost less and the existing low resolution imaging systems can still be utilized [1].

Super-resolution is based on the idea that a combination of low resolution (noisy) sequence of images of a scene can be used to generate a high resolution image or image sequence. Thus it attempts to reconstruct the original scene image with high resolution given a set of observed images at lower resolution. The general approach considers the low resolution images as resulting from resampling of a high resolution image. The goal is then to recover the high resolution image which when resampled based on the input images and the imaging model, will produce the low resolution observed images. Thus the accuracy of imaging model is vital for super-resolution and an incorrect modeling, say of motion, can actually degrade the image further [1]. The observed images could be taken from one or multiple cameras or could be frames of a video sequence. These images need to be mapped to a common reference frame. This process is registration. The super-resolution procedure can then be applied to a region of interest in the aligned composite image. The key to successful super-resolution consists of accurate alignment i.e. registration and formulation of an appropriate forward image model [1].

Typical super-resolution algorithms based on wavelets produces high-resolution (HR) image from a set of low-resolution (LR) frames. The relative motions in successive frames are estimated and used for aligning: HR image reconstruction from the set of LR images by performing image registration and then wavelet super-resolution [2-6].

The sample points in each frame into a HR grid. There are various types of models [7], [8] used to represent camera motion, namely, translation, rigid, affine, bilinear, and projective. The most general model is the projective model which has eight motion parameters. After registering all LR frames into a HR grid, the available samples distribute nonuniformly. This irregular sampling is called interlaced sampling. Then the wavelet super-resolution algorithm will be applied in order to get the HR image.

Methods for super-resolution can be broadly classified into three families of methods:
*i)* the classical multi-image super-resolution (combining images obtained at subpixel misalignments) [1-9],
*ii)* example-based super-resolution (learning correspondence between low and high resolution image patches from a database) [10-24], and
*iii)* an unified framework for combining these two families of methods. [25].

Other classification is based on:
*a)* multi-image super-resolution [1-9], and
*b)* still-image super-resolution [10-25].

In this paper we propose a new still-image super-resolution method (in the wavelet domain) based on the reconstruction of missing details subbands pixels at a given *ith* level via Rule of Three (Ro3) between pixels of approximation subband of such level, and pixels of approximation and detail subbands of (*i+1*)*th* level.

The Bidimensional Discrete Wavelet Transform and the method to reduce noise and to compress by wavelet thresholding is outlined in Section II. The new super-resolution method based on Ro3 is presented in Section III. The new approach as denoiser and compression tools in wavelet domain is outlined in Section IV. In Section V, we discuss briefly the more appropriate metrics for denoising and compression. In Section VI, the experimental results using the proposed algorithm are presented. Finally, Section VII provides a conclusion of the paper.

II. BIDIMENSIONAL DISCRETE WAVELET TRANSFORM

The Bidimensional Discrete Wavelet Transform (DWT-2D) [26, 27], [28-67] corresponds to multi-resolution approximation expressions. In practice, mutiresolution analysis is carried out using 4 channel filter banks (for each level of decomposition) composed of a low-pass and a high-pass filter and each filter bank is then sampled at a half rate (1/2 down sampling) of the previous frequency. By repeating this procedure, it is possible to obtain wavelet transform of any order. The down sampling procedure keeps the scaling parameter constant (equal to ½) throughout successive wavelet transforms so that is benefits for simple computer implementation. In the case of an image, the filtering is implemented in a separable way be filtering the lines and columns.

Note that [26, 27] the DWT of an image consists of four frequency channels for each level of decomposition. For example, for *i*-level of decomposition we have:

$LL_{n,i}$: Noisy Coefficients of Approximation.
$LH_{n,i}$: Noisy Coefficients of Vertical Detail,
$HL_{n,i}$: Noisy Coefficients of Horizontal Detail, and
$HH_{n,i}$: Noisy Coefficients of Diagonal Detail.

The LL part at each scale is decomposed recursively, as illustrated in Fig. 1 [26, 27].

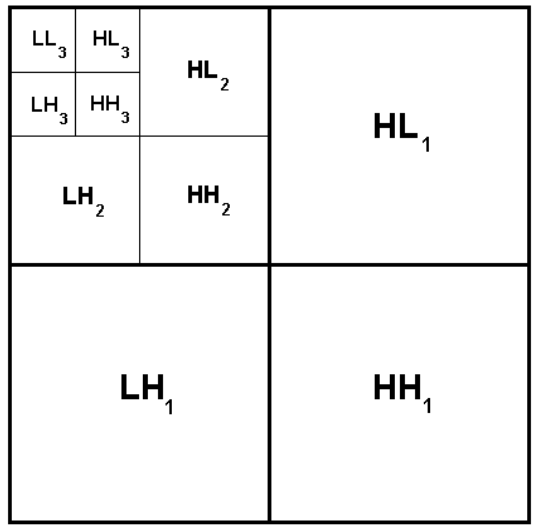

Fig. 1 Data preparation of the image. Recursive decomposition of LL parts.

To achieve space-scale adaptive noise reduction, we need to prepare the 1-D coefficient data stream which contains the space-scale information of 2-D images. This is somewhat similar to the "zigzag" arrangement of the DCT (Discrete Cosine Transform) coefficients in image coding applications [58]. In this data preparation step, the DWT-2D coefficients are rearranged as a 1-D coefficient series in spatial order so that the adjacent samples represent the same local areas in the original image [60].

Figure 2 shows the interior of the DWT-2D with the four subbands of the transformed image [67], which will be used in Fig.3. Each output of Fig. 2 represents a subband of splitting process of the 2-D coefficient matrix corresponding to Fig. 1.

### A. Wavelet Noise Thresholding

The wavelet coefficients calculated by a wavelet transform represent change in the image at a particular resolution. By looking at the image in various resolutions it should be possible to filter out noise, at least in theory. However, the definition of noise is a difficult one. In fact, "one person's noise is another's signal". In part this depends on the resolution one is looking at. One algorithm to remove Gaussian white noise is summarized by D. L. Donoho and I. M. Johnstone [68, 69], and synthesized in Fig. 3.

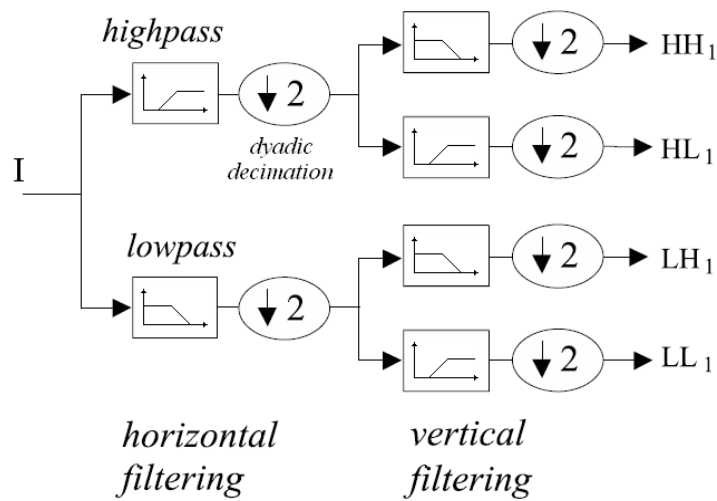

Fig. 2 Two dimensional DWT. A decomposition step. Usual splitting of the subbands.

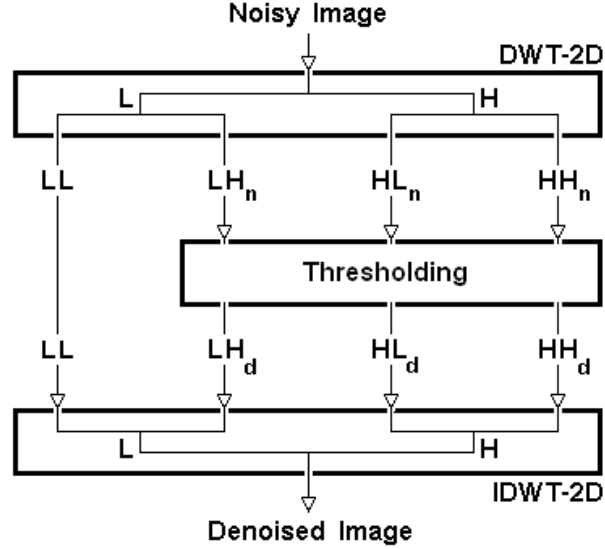

Fig. 3 Thresholding Techniques

The algorithm is:

1) Calculate a wavelet transform and order the coefficients by increasing frequency. This will result in an array containing the image average plus a set of coefficients of length 1, 2, 4, 8, etc. The noise threshold will be calculated on the highest frequency coefficient spectrum (this is the largest spectrum).

2) Calculate the *median absolute deviation* (mad) on the largest coefficient spectrum. The median is calculated from the absolute value of the coefficients. The equation for the median absolute deviation is shown below:

$$\delta_{mad} = \frac{median(|C_{n,i}|)}{0.6745} \qquad (1)$$

where $C_{n,i}$ may be $LH_{n,i}$, $HL_{n,i}$, or $HH_{n,i}$ for *i*-level of decomposition. The factor 0.6745 in the denominator rescales the numerator so that $\delta_{mad}$ is also a suitable estimator for the standard deviation for Gaussian white noise [70], [58], [60].

3) For calculating the noise threshold $\lambda$ we have used a modified version of the equation that has been discussed in papers by D. L. Donoho and I. M. Johnstone. The equation is:

$$\lambda = \delta_{mad}\sqrt{2log[N]} \qquad (2)$$

where *N* is the number of pixels in the subimage, i.e., HL, LH or HH.

4) Apply a thresholding algorithm to the coefficients. There are two popular versions:

**4.1. Hard thresholding**. Hard thresholding sets any coefficient less than or equal to the threshold to zero, see Fig. 4(a), where *x* may be $LH_{n,i}$, $HL_{n,i}$, or $HH_{n,i}$, *y* may be $HH_{d,i}$:
Denoised Coefficients of Diagonal Detail,
$HL_{d,i}$: Denoised Coefficients of Horizontal Detail,
$LH_{d,i}$: Denoised Coefficients of Vertical Detail,
for *i*-level of decomposition.

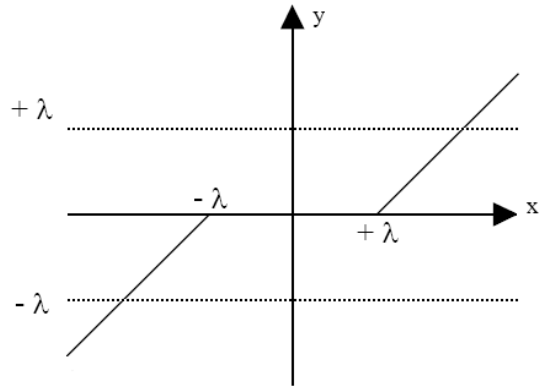

Fig. 4(a) Soft-Thresholfing

**The respective code is:**

```
for row = 1:N^(1/2)
  for column = 1:N^(1/2)
    if |C_{n,i}[row][column]| <= λ,
      C_{n,i}[row][column] = 0.0;
    end
  end
end
```

**4.2. Soft thresholding**. Soft thresholding sets any coefficient less than or equal to the threshold to zero, see Fig. 4(b). The threshold is subtracted from any coefficient that is greater than the threshold. This moves the image coefficients toward zero.

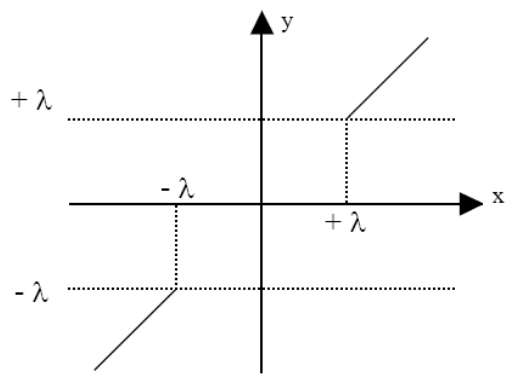

Fig. 4(b): Hard-Thresholfing

**The respective code is:**

```
for row = 1:N^(1/2)
  for column = 1:N^(1/2)
    if |C_{n,i}[row][column]| <= λ,
      C_{n,i}[row][column] = 0.0;
    else
      C_{n,i}[row][column] = C_{n,i}[row][column] - λ;
    end
  end
end
```

III. Rule Of Three Superresolution Method

In mathematics, specifically in elementary arithmetic and elementary algebra, given an equation between two fractions or rational expressions, one can cross-multiply to simplify the equation or determine the value of a variable [71].

Given an equation like:

$$\frac{a}{b} = \frac{c}{d} \qquad (3)$$

where $b$ and $d$ are not zero, one can cross-multiply to get:

$$ad = bc \quad \text{or} \quad a = \frac{bc}{d} \qquad (4)$$

In Euclidean geometry the same calculation can be achieved by considering the ratios as those of similar triangles.

Ro3 (this was sometimes also referred to as the Golden Rule, though that usage is rare compared to other uses of Golden Rule [72]) was a shorthand version for a particular form of cross-multiplication, often taught to students by rote. This rule was already known to Hebrews by the 15th century BC as it is a special case of the *Kal va-chomer*. It was also known by Indian (Vedic) mathematicians in the 6th century BC and Chinese mathematicians prior to the 7th century BC [73], though it was not used in Europe until much later [71]. Ro3 gained notoriety for being particularly difficult to explain: see Cocker's Arithmetick [74] for an example of how the premier textbook in the 17th century approached the subject.

For an equation of the form:

$$\frac{a}{b} = \frac{c}{x} \qquad (5)$$

where the variable to be evaluated is in the right-hand denominator, Ro3 states that:

$$x = \frac{bc}{a} \qquad (6)$$

In this context, $a$ is referred to as the extreme of the proportion, and $b$ and $c$ are called the means.

In short, Ro3 is a single conservation principle of proportionality between two ratios.

That principle becomes important in the wavelet domain by comparing the histogramic profiles between related subbands of consecutive resolutions. For example, Fig.5 an 6 shows this principle, where if $LL^0$ is the low-resolution original image $I_{LR}$, applying DWT-2D on it, we obtain four subbands $[LL^1, LH^1, HL^1, HH^1]$. Then, we apply Ro3 between those consecutive levels of splitting, as follows,

$$\frac{LL^1_{1,1}}{LL^0_{1,2}} = \frac{LH^1_{1,1}}{\hat{LH}^0_{1,2}} \qquad (7)$$

where the variable to be evaluated is in the right-hand denominator, Ro3 states that:

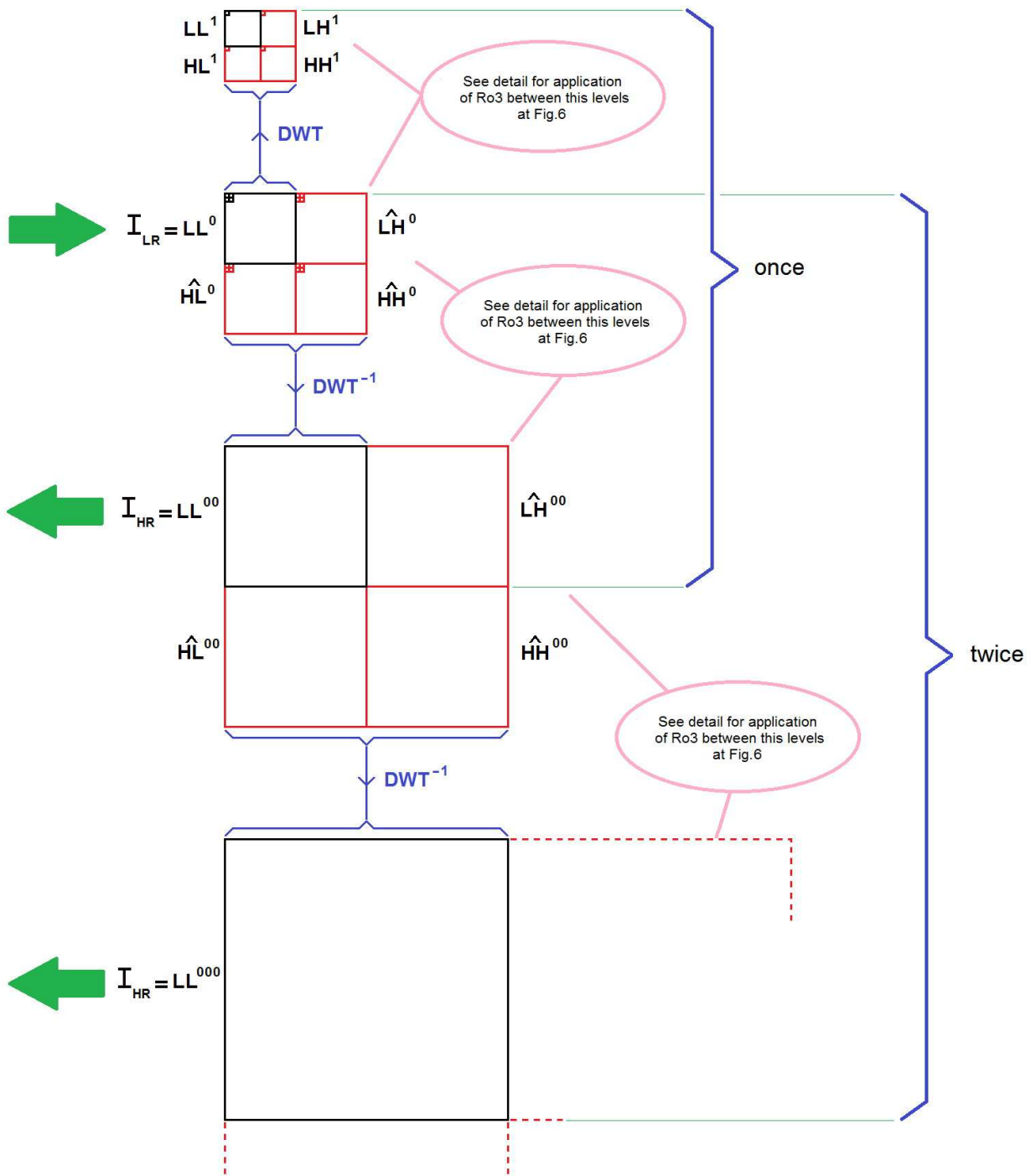

Fig. 5: Application of Ro3 between several couple of levels.

$$\hat{L}H^0_{1,2} = \frac{LH^1_{1,1} LL^0_{1,2}}{LL^1_{1,1}} \tag{8}$$

Figure 6 shows in detail this procedure for twelve examples, relatives to Eq.(8). Histogramic profiles (between respective subbands of consecutive levels of splitting) of Fig.7 are witnesses of these statements. Specifically, Fig.8 shows the comparative histogramic profiles between $I_{HR} = LL^{00}$, $I_{LR} = LL^0$ and $LL^1$, which indicates the consistency of the employed criteria.

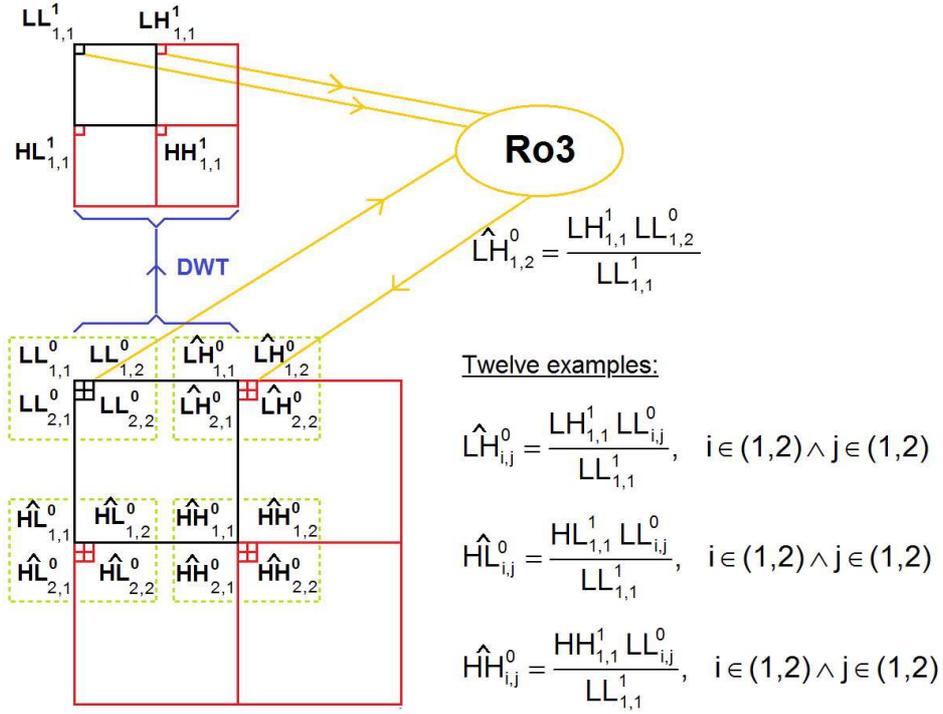

Fig. 6: Detail for application of Ro3 between two consecutive levels of splitting.

According to Fig.5 the procedure for superresolution configuration is the follow:

*once*:

- Input $LL^0 = I_{LR} \in \Re^{ROW \times COL}$

- $[LL^1, LH^1, HL^1, HH^1] = DWT\text{-}2D(LL^0)$

- $\hat{LH}^0_{r,c} = Ro3(LH^1_{r/2,c/2}, LL^0_{r,c}, LL^1_{r/2,c/2}) = \dfrac{LH^1_{r/2,c/2} LL^0_{r,c}}{LL^1_{r/2,c/2}} \quad \forall r,c$

- $\hat{HL}^0_{r,c} = Ro3(HL^1_{r/2,c/2}, LL^0_{r,c}, LL^1_{r/2,c/2}) = \dfrac{HL^1_{r/2,c/2} LL^0_{r,c}}{LL^1_{r/2,c/2}} \quad \forall r,c$

- $\hat{HH}^0_{r,c} = Ro3(HH^1_{r/2,c/2}, LL^0_{r,c}, LL^1_{r/2,c/2}) = \dfrac{HH^1_{r/2,c/2} LL^0_{r,c}}{LL^1_{r/2,c/2}} \quad \forall r,c$

- $LL^{00} = DWT\text{-}2D^{-1}(LL^0, \hat{LH}^0, \hat{HL}^0, \hat{HH}^0)$

- Output $I_{HR} = LL^{00}$ (first high-resolution resulting image) $\in \Re^{(2 \times ROW) \times (2 \times COL)}$

*twice*:

- Input $LL^{00} = I_{HR} \in \Re^{(2 \times ROW) \times (2 \times COL)}$

- $[LL^0, \hat{LH}^0, \hat{HL}^0, \hat{HH}^0] = DWT\text{-}2D(LL^{00})$

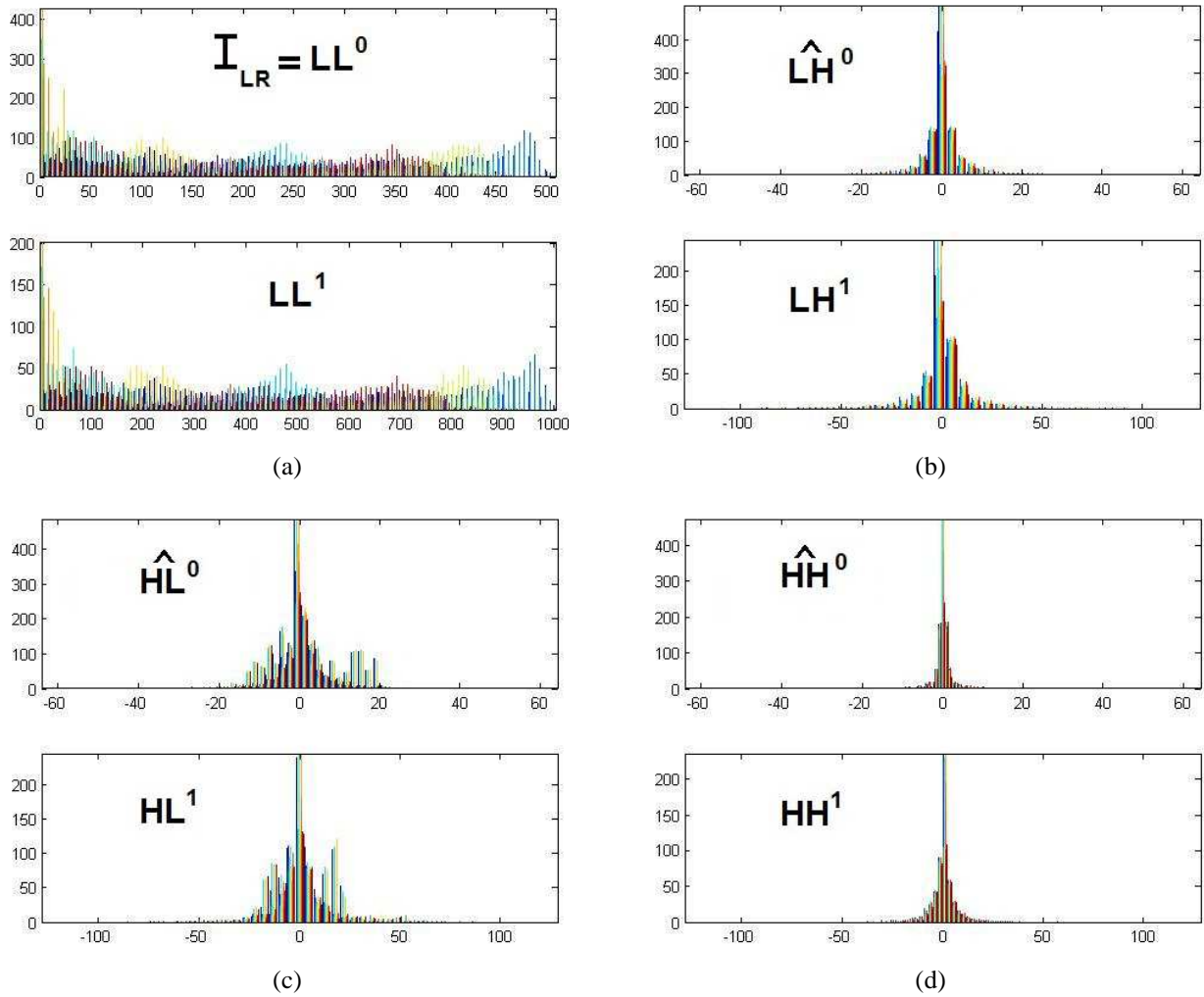

Fig. 7: Histogramic profiles between respective subbands of consecutive levels of splitting.

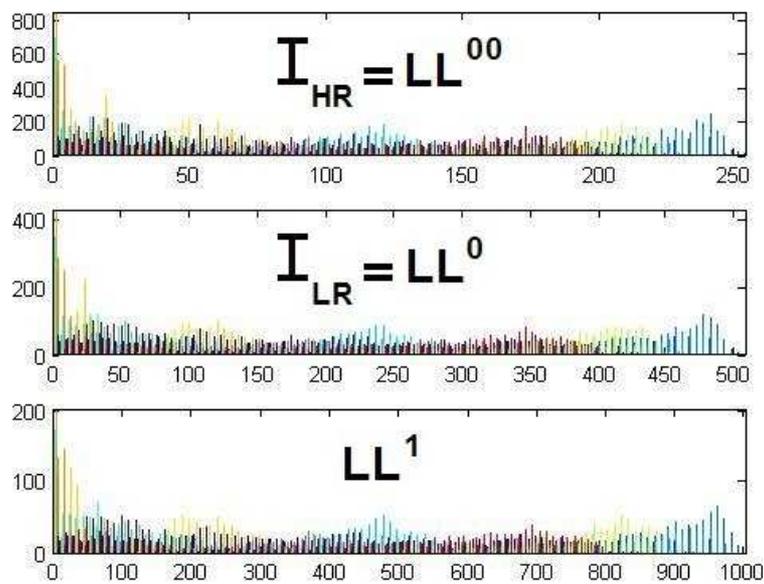

Fig. 8: Comparative histogramic profiles between $I_{HR} = LL^{00}$, $I_{LR} = LL^{0}$ and $LL^{1}$.

- $\hat{LH}_{r,c}^{00} = Ro3(\hat{LH}_{r/2,c/2}^{0}, LL_{r,c}^{00}, LL_{r/2,c/2}^{0}) = \dfrac{\hat{LH}_{r/2,c/2}^{0} LL_{r,c}^{00}}{LL_{r/2,c/2}^{0}} \quad \forall 2r, 2c$

$\hat{HL}_{r,c}^{00} = Ro3(\hat{HL}_{r/2,c/2}^{0}, LL_{r,c}^{00}, LL_{r/2,c/2}^{0}) = \dfrac{\hat{HL}_{r/2,c/2}^{0} LL_{r,c}^{00}}{LL_{r/2,c/2}^{0}} \quad \forall 2r, 2c$

$\hat{HH}_{r,c}^{00} = Ro3(\hat{HH}_{r/2,c/2}^{0}, LL_{r,c}^{00}, LL_{r/2,c/2}^{0}) = \dfrac{\hat{HH}_{r/2,c/2}^{0} LL_{r,c}^{00}}{LL_{r/2,c/2}^{0}} \quad \forall 2r, 2c$

- $LL^{000} = DWT\text{-}2D^{-1}(LL^{00}, \hat{LH}^{00}, \hat{HL}^{00}, \hat{HH}^{00})$

- Output $I_{HR} = LL^{000}$ (second high-resolution resulting image) $\in \Re^{(4xROW)x(4xCOL)}$

IV. COMPRESSION AND DENOISING VIA RO3

An image is affected by noise in its acquisition and processing. The denoising techniques are used to remove the additive noise while retaining as much as possible the important image features. In the recent years there has been an important amount of research on wavelet thresholding and threshold selection for images denoising [75, 76], because wavelet provides an appropriate basis for separating noisy signal from the image signal. The motivation is that as the wavelet transform is good at energy compaction, the small coefficients are more likely due to noise and large coefficient due to important signal features [76]. These small coefficients can be thresholded without affecting the significant features of the image.

In fact, the thresholding technique is the last approach based on wavelet theory to provide an enhanced approach for eliminating such noise source and ensure better image quality [76]. Thresholding is a simple non-linear technique, which operates on one wavelet coefficient at a time. In its basic form, each coefficient is thresholded by comparing against threshold, if the coefficient is smaller than threshold, set to zero; otherwise it is kept or modified. Replacing the small noisy coefficients by zero and inverse wavelet transform on the result may lead to reconstruction with the essential signal characteristics and with less noise. Since the work of Donoho & Johnstone [68, 69], there has been much research on finding thresholds, however few are specifically designed for images [76].

Unfortunately, this technique has the following disadvantages:
1) it depends on the correct election of the type of thresholding, e.g., OracleShrink, VisuShrink (soft-thresholding, hard-thresholding, and semi-soft-thresholding), Sure Shrink, Bayesian soft thresholding, Bayesian MMSE estimation, Thresholding Neural Network (TNN), due to Zhang, NormalShrink, , etc. [76]
2) it depends on the correct estimation of the threshold which is arguably the most important design parameter,
3) it doesn't have a fine adjustment of the threshold after their calculation,
4) it should be applied at each level of decomposition, needing several levels, and
5) the specific distributions of the signal and noise may not be well matched at different scales.

Therefore, a new method without these constraints will represent an upgrade. On the other hand, similar considerations should be kept in mind regarding the problem of image compression based on wavelet thresholding.

Figure 9 shows us Ro3 applied to image compression and denoising, which is clearly different to the traditional methods of denoising and compression in wavelet domain [75], including despeckling, that is to say, the suppression of multiplicative noise of Gamma distribution in Synthetic Aperture Radar (SAR) imagery [76]. On the other hand, both Ro3 and the traditional methods of denoising and compression in wavelet domain, compression and denoising mean the same thing. Finally, whole method is based on the fact that resolutions related of subsequent bands have the same histogramic profile, which can be clearly seen in Fig.10, where $I$, is the original image, $\hat{I}$ is the denoised (compressed/decompressed) image and $\tilde{I}$ is the deblurred image.

Eventually, and if the original image is small, i.e., less than 512-by-512 pixels, we employ a deblurring mask (for edge enhancement) after decompression (at the end of denoising procedure). Said mask parameters are calculated using a genetic algorithm [77-79]. Later, we can see that mask in MATLAB® code [80].

Main application of this technology is for removal the detail subbands of image (in wavelet domain) used in cellular phones (mobile platforms) in order to save bandwidth in the transmission thereof, and memory for storage [81, 82]. In this way, we can increase the compression between 4 and 14 times more than that usually available in such mobile devices. In fact, Ro3 acts as a compression catalyst for the true compression algorithm compresses more. The compression catalyst concept already well established in the literature [77-79].

**NOTE:** In all figures shown here (including Fig.9), Ro3 is the procedure that exclusively represents the rule of three, however, from now on, as well as in subsequent MATLAB® sources, Ro3 represents a downsampling process via DWT-2D (inside encoder) and Ro3$^{-1}$ represents a downsampling process via DWT-2D, the rule of three itself, and an upsampling process via DWT-2D$^{-1}$ (inside decoder).

*Encoder:*
- Enter the original image (raw-data) $\mathbf{I} = \mathbf{LL^0}$
- <u>Ro3</u>: DWT-2D on $\mathbf{LL^0}$ (compression via downsampling). We obtain [$\mathbf{LL^1}$, $\mathbf{LH^1}$, $\mathbf{HL^1}$, $\mathbf{HH^1}$].
    We discard $\mathbf{LH^1}$, $\mathbf{HL^1}$ and $\mathbf{HH^1}$
- We coded $\mathbf{LL^1}$ with JPEG or JPEG2000 [83,84]

*Decoder (without deblurring):*
- We decoded $\mathbf{LL^1}$ with JPEG$^{-1}$ or JPEG2000$^{-1}$
- <u>Ro3$^{-1}$</u>: DWT-2D on $\mathbf{LL^1}$. We obtain [$\mathbf{LL^2}$, $\mathbf{LH^2}$, $\mathbf{HL^2}$, $\mathbf{HH^2}$].
    With $\mathbf{LL^1}$ and $\mathbf{LL^2}$, $\mathbf{LH^2}$, $\mathbf{HL^2}$, $\mathbf{HH^2}$, we obtain $\mathbf{\hat{L}H^1}$, $\mathbf{\hat{H}L^1}$ and $\mathbf{\hat{H}H^1}$.
    DWT-2D$^{-1}$ on [$\mathbf{LL^1}$, $\mathbf{\hat{L}H^1}$, $\mathbf{\hat{H}L^1}$, $\mathbf{\hat{H}H^1}$]. We obtain $\mathbf{\hat{L}L^0}$.
- $\mathbf{\hat{I}} = \mathbf{\hat{L}L^0}$

*Decoder (with deblurring):*
- We decoded $\mathbf{LL^1}$ with JPEG$^{-1}$ or JPEG2000$^{-1}$
- <u>Ro3$^{-1}$</u>: DWT-2D on $\mathbf{LL^1}$. We obtain [$\mathbf{LL^2}$, $\mathbf{LH^2}$, $\mathbf{HL^2}$, $\mathbf{HH^2}$].
    With $\mathbf{LL^1}$ and $\mathbf{LL^2}$, $\mathbf{LH^2}$, $\mathbf{HL^2}$, $\mathbf{HH^2}$, we obtain $\mathbf{\hat{L}H^1}$, $\mathbf{\hat{H}L^1}$ and $\mathbf{\hat{H}H^1}$.
    DWT-2D$^{-1}$ on [$\mathbf{LL^1}$, $\mathbf{\hat{L}H^1}$, $\mathbf{\hat{H}L^1}$, $\mathbf{\hat{H}H^1}$]. We obtain $\mathbf{\hat{L}L^0}$.
- $\mathbf{\hat{I}} = \mathbf{\hat{L}L^0}$
- $\mathbf{\tilde{I}} = \mathbf{deblurring}\,(\mathbf{\hat{I}})$

On the other hand, if we need to perform a superresolution on an image, the procedure shall be as specified as decoder. Instead, if we need to do a compression or denoising on an image, then the procedure is a combination of encoder and decoder.

We present below a series of MATLAB® routines organized according to their functionality, i.e., first, the encoder, which calls to function Ro3 fundamentally, and secondarily to built-in MATLAB® functions.

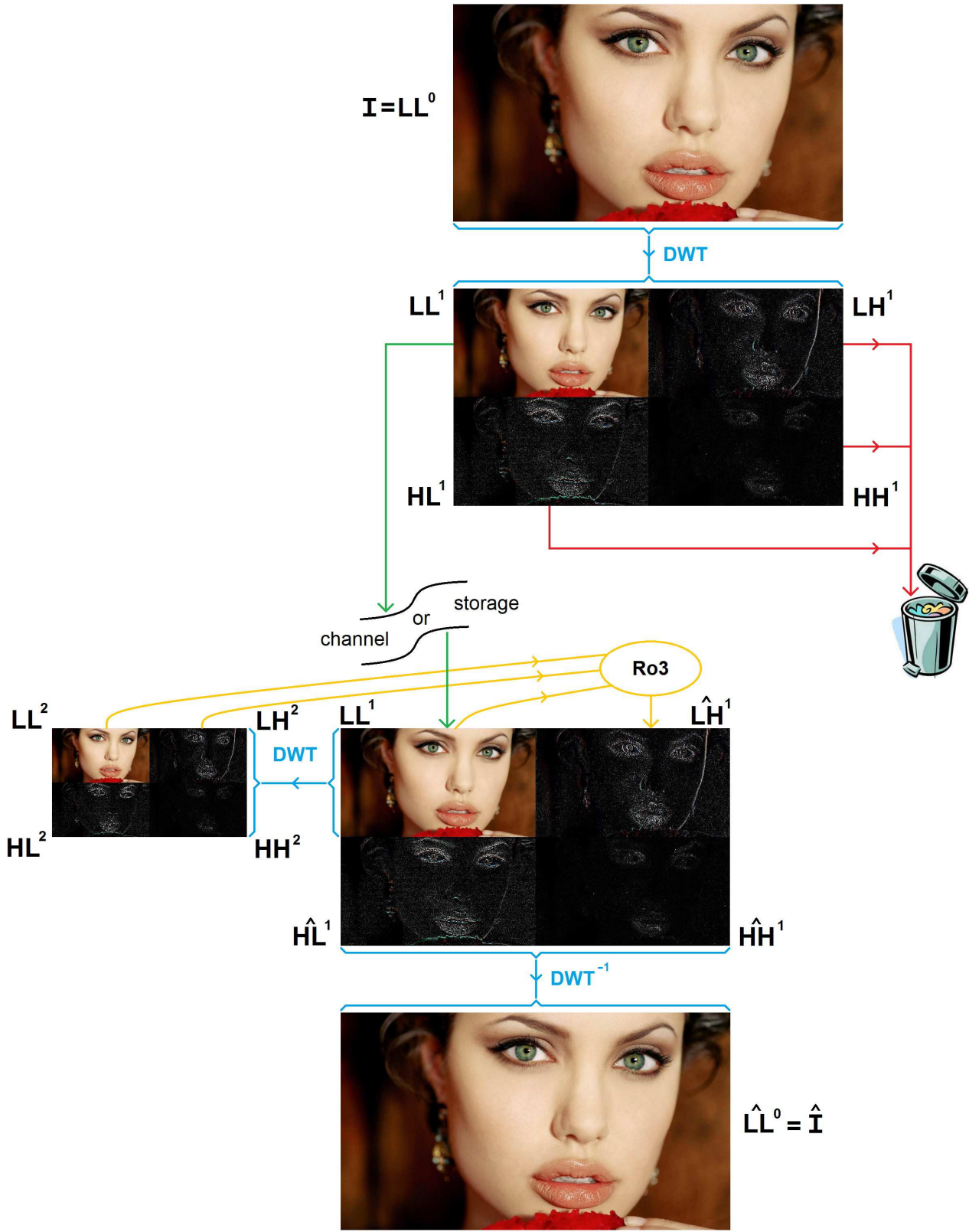

Fig. 9: Ro3 applied to image compression and denoising.

Later, we present the decoder, which calls mainly to function iRo3, and built-in MATLAB® functions too. Finally, we incorporate the deblurring functions.

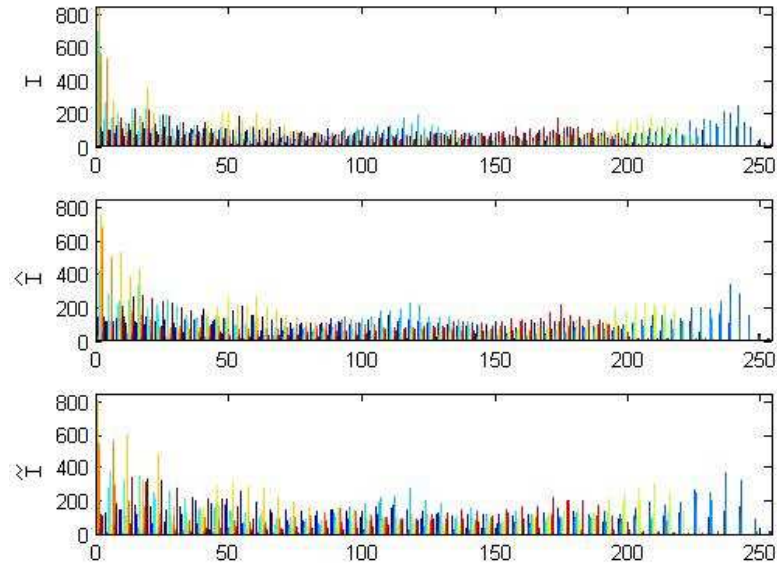
Fig. 10: Comparative histogramic profiles between $I$, $\hat{I}$ and $\tilde{I}$.

Inside iRo3 code, we can see in color (red for LH1, green for HL1 and blue for HH1) a parameter, known as, anchoring parameter (ap << 1), which is useful in cases where LL2 represents a subband with very dark pixels, i.e., very close to zero pixels, in which cases, and the absence of the anchoring parameter greater than zero, the ratio is undefined.

**MATLAB's code for encoder:**

```
format long g
namefile = input('namefile: ','s');
nemafile2 = strcat(namefile,'.jpg');
I = imread(namefile2);
I = single(I);

[rI,cI,bI] = size(I);

basis = input('wavelet basis: ','s'); % suggested db1
LL1 = [];
for b = 1:bI
   LL1(:,:,b) = Ro3(I(:,:,b),basis);
end
clear I;
LL1 = LL1/2;

namefile3 = strcat(namefile,'c');
namefile3 = strcat(namefile3,'.jpg');
LL1 = uint8(LL1);
imwrite(LL1,namefile3)
```

**MATLAB's code for Ro3:**

```
function LL1 = Ro3(I,basis)

[LL1,LH1,HL1,HH1] = dwt2(I,basis,'mode','sym');
return;
```

### MATLAB's code for decoder:

```
format long g
namefile = input('namefile: ','s');
namefile2 = strcat(namefile,'c');
namefile2 = strcat(namefile2,'.jpg');
LL1 = imread(namefile2);
LL1 = single(LL1);

[rLL1,cLL1,bLL1] = size(LL1);

basis = input('wavelet basis: ','s'); % suggested db1
Ib = [];
for b = 1:bLL1
   Ib(:,:,b) = iRo3(LL1(:,:,b),basis);
end
clear LL1;

namefile3 = strcat(namefile,'b');
namefile3 = strcat(namefile3,'.jpg');
Ib = uint8(Ib);
imwrite(Ib,namefile3)

Id = deblurring(Ib);

namefile4 = strcat(namefile,'d');
namefile4 = strcat(namefile4,'.jpg');
Id = uint8(Id);
imwrite(Id,namefile4);
```

### MATLAB's code for iRo3:

```
function Ib = iRo3(LL1,basis)

[rLL1,cLL1] = size(LL1);

[LL2,LH2,HL2,HH2] = dwt2(LL1,basis,'mode','sym');
ap = .0001; % usually ap << 1
for r2 = 1:rLL1/2
  for c2 = 1:cLL1/2
    r1 = 2*r2;
    c1 = 2*c2;
    LH1(r1-1,c1-1) = LH2(r2,c2)*(ap+LL1(r1-1,c1-1))/(ap+LL2(r2,c2));
    LH1(r1-1,c1  ) = LH2(r2,c2)*(ap+LL1(r1-1,c1  ))/(ap+LL2(r2,c2));
    LH1(r1  ,c1-1) = LH2(r2,c2)*(ap+LL1(r1  ,c1-1))/(ap+LL2(r2,c2));
    LH1(r1  ,c1  ) = LH2(r2,c2)*(ap+LL1(r1  ,c1  ))/(ap+LL2(r2,c2));
    HL1(r1-1,c1-1) = HL2(r2,c2)*(ap+LL1(r1-1,c1-1))/(ap+LL2(r2,c2));
    HL1(r1-1,c1  ) = HL2(r2,c2)*(ap+LL1(r1-1,c1  ))/(ap+LL2(r2,c2));
    HL1(r1  ,c1-1) = HL2(r2,c2)*(ap+LL1(r1  ,c1-1))/(ap+LL2(r2,c2));
    HL1(r1  ,c1  ) = HL2(r2,c2)*(ap+LL1(r1  ,c1  ))/(ap+LL2(r2,c2));
    HH1(r1-1,c1-1) = HH2(r2,c2)*(ap+LL1(r1-1,c1-1))/(ap+LL2(r2,c2));
    HH1(r1-1,c1  ) = HH2(r2,c2)*(ap+LL1(r1-1,c1  ))/(ap+LL2(r2,c2));
    HH1(r1  ,c1-1) = HH2(r2,c2)*(ap+LL1(r1  ,c1-1))/(ap+LL2(r2,c2));
    HH1(r1  ,c1  ) = HH2(r2,c2)*(ap+LL1(r1  ,c1  ))/(ap+LL2(r2,c2));
  end
end
clear LL2, LH2, HL2, HH2;
LL1 = 2*LL1;
Ib = idwt2(LL1,LH1,HL1,HH1,basis,'mode','sym');
return;
```

**MATLAB's code for deblurring:**

```
function Id = deblurring(Ib)

[rIb,cIb,bIb] = size(Ib);
for b = 1:bIb
   Id(:,:,b) = ddeblurmp(Ib(:,:,b));
end

return;

%%%%%%%%%%%%%%%%%%%%%%

function Id = ddeblurmp(Ib)

[rIb,cIb] = size(Ib);
m = 7; d = floor(m/2);
Ibaux = zeros(rIb+2*d,cIb+2*d);
Ibaux(1+d:d+rIb,1+d:d+cIb) = Ib;
clear Ib;
ri = 1;
rf = rIb+2*d;
ci = 1;
cf = cIbL+2*d;
Ibaux(ri:rf,ci:cf) = deblurmp(Ibaux(ri:rf,ci:cf));
Id = Ibaux(1+d:d+rIb,1+d:d+cIb);

return;

%%%%%%%%%%%%%%%%%%%%%%

function Id = deblurmp(Ib)

[rIb,cIb] = size(Ib);
w = 7;
Ibaux = Ib; Id = Ib;

  % Convolución vertical:
  for r = 1+floor(w/2):rIb-floor(w/2)
    for c = 1:cIb
      acuv = 0;
      for rw = 1:w
        acuv = acuv + Ib(r-(1+floor(w/2))+rw,c)*(-0.0129);
      end
      Ibaux(r,c) = acuv;
    end
  end

  % Convolución horizontal:
  for c = 1+floor(w/2):cIb-floor(w/2)
    for r = 1+floor(w/2):rIb-floor(w/2)
      acuh = 0;
      for cw = 1:w
        acuh = acuh + Ibaux(r,c-(1+floor(w/2))+cw);
      end
      Id(r,c) = acuh + 1.63*Ib(r,c);
    end
  end

return;
```

## V. METRICS

### A. Data Compression Ratio (CR)

Data compression ratio, also known as compression power, is a computer-science term used to quantify the reduction in data-representation size produced by a data compression algorithm. The data compression ratio is analogous to the physical compression ratio used to measure physical compression of substances, and is defined in the same way, as the ratio between the *uncompressed size* and the *compressed size* [85]:

$$CR = \frac{Uncompressed\ Size}{Compressed\ Size} \qquad (3)$$

Thus a representation that compresses a 10MB file to 2MB has a compression ratio of 10/2 = 5, often notated as an explicit ratio, 5:1 (read "five to one"), or as an implicit ratio, 5X. Note that this formulation applies equally for compression, where the uncompressed size is that of the original; and for decompression, where the uncompressed size is that of the reproduction.

### B. Percent Space Savings (PSS)

Sometimes the space savings is given instead, which is defined as the reduction in size relative to the uncompressed size:

$$PSS = \left(1 - \frac{1}{CR}\right) * 100\% \qquad (4)$$

Thus a representation that compresses 10MB file to 2MB would yield a space savings of 1-2/10 = 0.8, often notated as a percentage, 80%.

### C. Peak Signal-To-Noise Ratio (PSNR)

The phrase peak signal-to-noise ratio, often abbreviated PSNR, is an engineering term for the ratio between the maximum possible power of a signal and the power of corrupting noise that affects the fidelity of its representation. Because many signals have a very wide dynamic range, PSNR is usually expressed in terms of the logarithmic decibel scale.

The PSNR is most commonly used as a measure of quality of reconstruction in image compression etc [85]. It is most easily defined via the mean squared error (MSE) which for two $NR \times NC$ (rows-by-columns) monochrome images $I$ and $I_d$, where the second one of the images is considered a decompressed/denoised approximation of the other is defined as:

$$MSE = \frac{1}{NRxNC} \sum_{nr=0}^{NR-1} \sum_{nc-0}^{NC-1} \left\| I(nr,nc) - I_d(nr,nc) \right\|^2 \qquad (5)$$

The PSNR is defined as [84]:

$$PSNR = 10 \log_{10}\left(\frac{MAX_I^2}{MSE}\right) = 20 \log_{10}\left(\frac{MAX_I}{\sqrt{MSE}}\right) \qquad (6)$$

Here, $MAX_i$ is the maximum pixel value of the image. When the pixels are represented using 8 bits per sample, this is 255. More generally, when samples are represented using linear pulse code modulation (PCM) with B bits per sample, maximum possible value of $MAX_i$ is $2^B-1$.

For color images with three red-green-blue (RGB) values per pixel, the definition of PSNR is the same except the MSE is the sum over all squared value differences divided by image size and by three [85].

Typical values for the PSNR in lossy image and video compression are between 30 and 50 dB, where higher is better.

Finally, a conspicuous metric for these cases is the mean absolute error, which is a quantity used to measure how close forecasts or predictions are to the eventual outcomes. The mean absolute error (MAE) is given by

$$MAE = \frac{1}{NRxNC} \sum_{nr=0}^{NR-1} \sum_{nc=0}^{NC-1} \left\| I(nr,nc) - I_d(nr,nc) \right\| \quad (7)$$

which for two *NR×NC* (rows-by-columns) monochrome images *I* and *$I_d$* , where the second one of the images is considered a decompressed/denoised approximation of the other of the first one.

## VI. EXPERIMENTAL RESULTS

The simulations demonstrate that the new technique improves the noise reduction and compression performances in wavelet domain to the maximum.

Here, we present two experimental results using an image of Angelina Jolie. Such image (raw-data) was converted to bitmap file format for their treatment [83, 84]. Figure 11 shows the original image, noisy (Gaussian white noise, with mean value = 0, and standard deviation = 0.02) and filtered images, with 1920-by-1080 pixels by 256 levels per colour bitmap matrix. Table I summarizes the assessment parameters vs. filtering techniques for Fig.11, where ST means Soft-Thresholding and HT means Hard-Thresholding. On the other hand, Fig.12 shows the original and compressed/decompressed images via ST, HT and the new techniques acting as compression catalyst on JPEG and JPEG2000 compression formats, with and without deblurring. Table II summarizes the assessment parameters vs. compressed techniques for Fig.12. The quality is similar with very different CR and PSS. In all cases, Ro3 was applied only once.

TABLE I
ORIGINAL VS DENOISED IMAGES

| METRIC | ST | HT | RO3 | RO3+DEBL |
|---|---|---|---|---|
| MAE | 1.1770 | 1.0762 | 1.0865 | 1.6559 |
| MSE | 9.0604 | 8.5520 | 9.7704 | 15.9035 |
| PSNR | 38.5593 | 38.8101 | 38.2317 | 36.1159 |

TABLE II
ORIGINAL VS DECOMPRESSED IMAGES

| METRIC | JPEG | JPEG2K | JPEG+RO3 | JPEG+RO3+DEBL | JPEG2K+RO3 | JPEG2K+RO3+DEBL |
|---|---|---|---|---|---|---|
| CR | 35.0322 | 8.6120 | 138.5182 | | 24.0412 | |
| PSS [%] | 97.1454 | 88.3882 | 99.2780 | | 95.8404 | |
| MAE | 0.0510 | 0.0458 | 0.4190 | 0.5747 | 0.2889 | 0.5179 |
| MSE | 0.0855 | 0.0486 | 2.3188 | 3.6617 | 1.3345 | 3.1318 |
| PSNR | 58.8090 | 61.2636 | 44.4781 | 42.4939 | 46.8775 | 43.1728 |

Table II shows Ro3 as compression catalyst accompanying, first to JPEG, and second to JPEG2K. In both cases, the compression rate increases dramatically, while the quality assessment metrics remain unchanged.

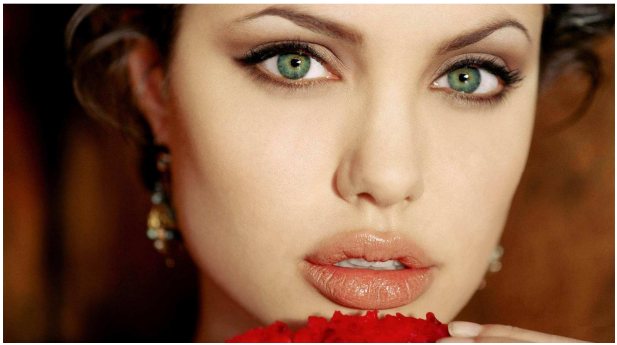
Original

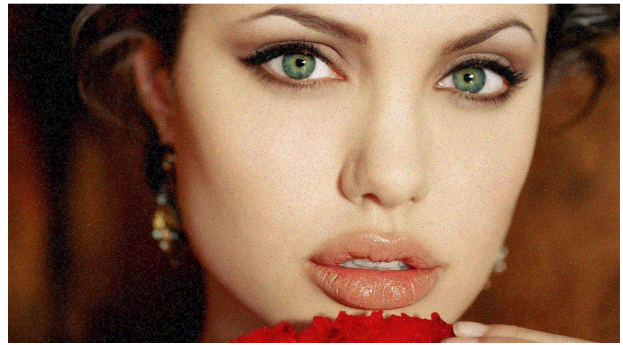
Noised

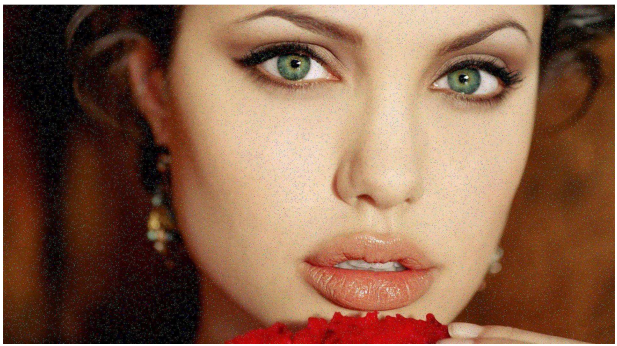
Soft-thresholding

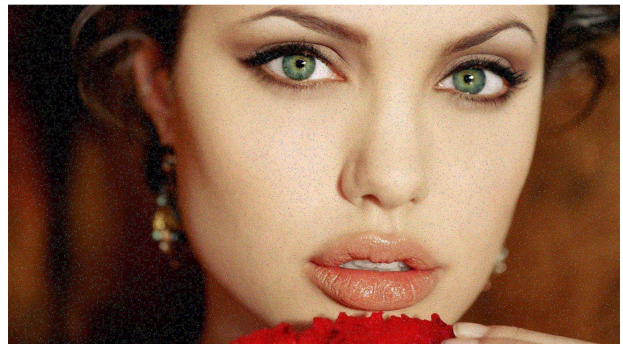
Hard-thresholding

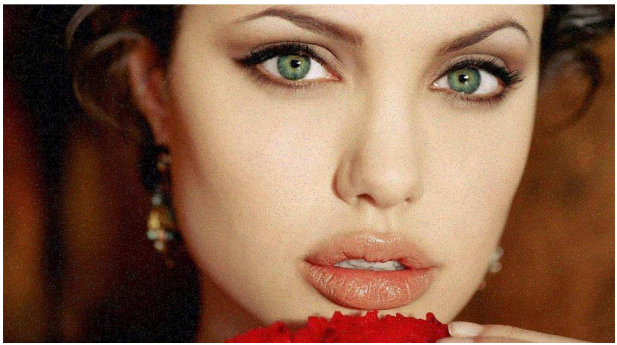
Ro3

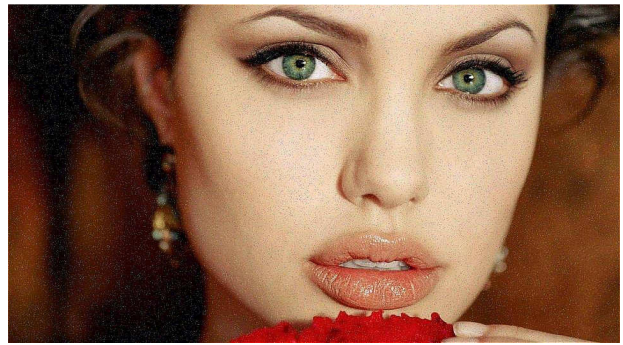
Ro3+Deblurring

Fig. 11: Denoising experiment.

On the other hand, Fig.10 shows the real histograms of $I$, $\hat{I}$ and $\tilde{I}$ for these experiments. It is clear that the histogramic profiles are quite similar. Similar case occurs in Fig.7 and 8, where we can see the histogramic affinity between subbands of subsequent levels (each in its respective level, i.e., resolution).

It is important to note, that there is a dramatic difference between the two experiments, that is, between denoising and compression. While in Table I the metrics give practically the same for all techniques, except when applied deblurring (which is logical in a noisy context), in Table II quality metric falls substantially when Ro3 applies and when it does not apply. However, metrics are excellent still, differences are imperceptible to the human eye and the respective compression rates are multiplied (at worst) by 4 for JPEG and JPEG2K either.

Wavelet basis employed in the experiments for Ro3 were Daubechies 1 (that is to say, Haar basis) with only one level of decomposition, while in the case of ST and HT was Daubechies 4.

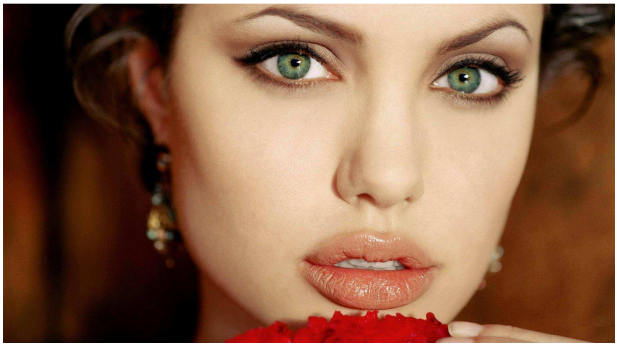
JPEG

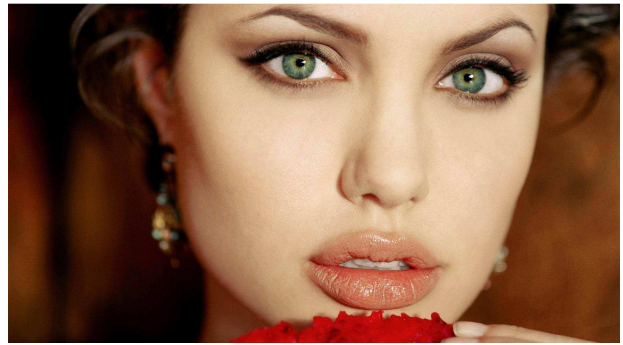
JPEG2K

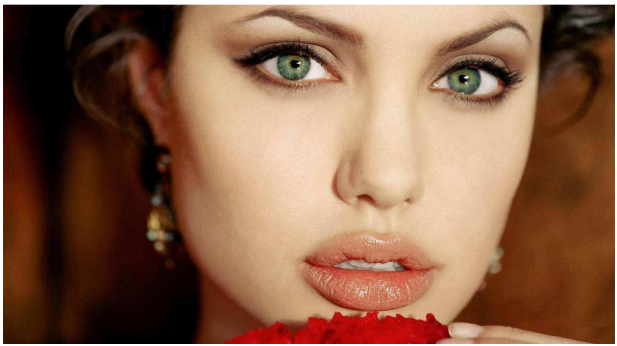
JPEG+R03

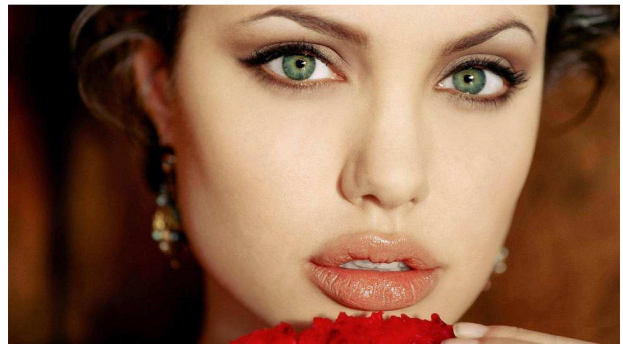
JPEG2K+Ro3

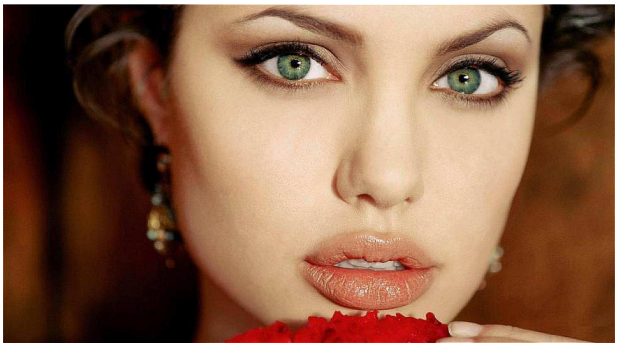
JPEG+Ro3+Deblurring

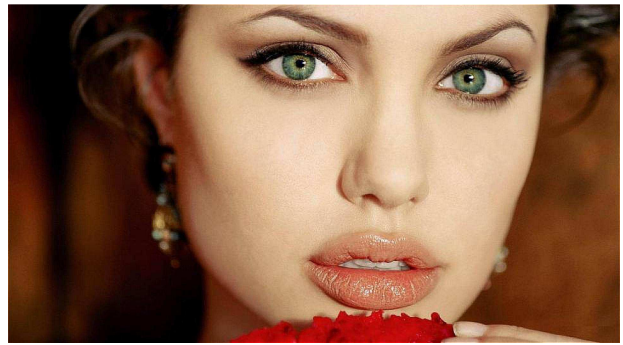
JPEG2K+Ro3+Deblurring

Fig. 12: Compression Experiment.

Finally, all techniques (superresolution, denoising and compression) were implemented in MATLAB® R2014a (Mathworks, Natick, MA) [85] on an Notebook with Intel® Core(TM) i5 CPU M 430 @ 2.27 GHz and 6 GB RAM on Microsoft® Windows 7© Home Premium 32 bits.

## VII. CONCLUSIONS

In this paper we have developed a *Rule of Three (Ro3)* technique for image superresolution, filtering and compression inside wavelet domain. The simulations show that the new have better performance than the most commonly used thresholding technique for compression and denoising (for the studied benchmark parameters) which include Soft-Thresholding and Hard-Thresholding.

Besides, the novel demonstrated to be efficient to remove multiplied noise, and all uncle of noise in the undecimated wavelet domain. Finally, cleaner images suggest potential improvements for classification and recognition.

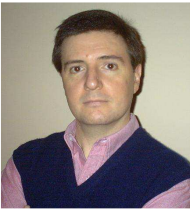 **Mario Mastriani** received the B.Eng. degree in 1989 and the Ph.D. degree in 2006, both in electrical engineering. Besides, he received the Ph.D. degree in Computer Sciences in 2009 and the Ph.D. degree in Science and Technology in 2011. He was the director of research and development laboratories in signal and image processing, technical innovation, and new technologies in several institutions. Currently, he is a CoFounder and the Chief Science Officer (CSO) of DLQS LLC, 4431 NW 63RD Drive, Coconut Creek, FL 33073, USA.. He published 42 papers. He was a reviewer of IEEE Transactions on Neural Networks, Signal Processing Letters, Transactions on Image Processing, Transactions on Signal Processing, Communications Letters, Transactions on Geoscience and Remote Sensing, Transactions on Medical Imaging, Transactions on Biomedical Engineering, Transactions on Fuzzy Systems, Transactions on Multimedia; Springer-Verlag Journal of Digital Imaging, SPIE Optical Engineering Journal; and Taylor & Francis International Journal of Remote Sensing. He was a member of IEEE, Piscataway, USA, during 9 years. He became a member of WASET in 2004. His areas of interest include Quantum Signal and Image Processing, and cutting-edge technologies for computing and communication.